\PassOptionsToPackage{table}{xcolor}
\documentclass[10pt,twocolumn,letterpaper]{article}

\usepackage[pagenumbers]{cvpr} 

\usepackage{url}            
\usepackage{booktabs}       
\usepackage{amsfonts}       
\usepackage{nicefrac}       
\usepackage{microtype}      
\usepackage{graphicx}
\usepackage{algorithm}
\usepackage{algorithmicx}
\usepackage{algpseudocode}
\usepackage{amsmath}
\usepackage{tabularx}
\usepackage{subcaption}
\newfloat{figtab}{htb}{fgtb}
\makeatletter
  \newcommand\figcaption{\def\@captype{figure}\caption}
  \newcommand\tabcaption{\def\@captype{table}\caption}
\makeatother
\usepackage{multirow}
\usepackage{wrapfig}
\usepackage{amssymb}
\usepackage{pifont}
\usepackage{threeparttable}
\usepackage{makecell}

\definecolor{cvprblue}{rgb}{0.21,0.49,0.74}
\usepackage[pagebackref,breaklinks,colorlinks,allcolors=cvprblue]{hyperref}

\def\paperID{3600} 
\def\confName{CVPR}
\def\confYear{2025}

\title{AR-Diffusion: Asynchronous Video Generation with Auto-Regressive Diffusion}

\author{Mingzhen Sun*\\
IA, CAS\\
UCAS\\
\and
Weining Wang*\\
IA, CAS\\
UCAS\\
\and
Gen Li\\
Bytedance Inc.\\
\and
Jiawei Liu\\
Bytedance Inc.\\
\and
Jiahui Sun\\
IA, CAS\\
UCAS\\
\and
Wanquan Feng\\
Bytedance Inc.\\
\and
Shanshan Lao\\
Bytedance Inc.\\
\and
Siyu Zhou\\
Bytedance Inc.\\
\and
Qian He\\
Bytedance Inc.\\
\and
Jing Liu\\
IA, CAS\\
UCAS\\
}

\PassOptionsToPackage{table}{xcolor}
\begin{document}

\maketitle
\begin{abstract}
The task of video generation requires synthesizing visually realistic and temporally coherent video frames. Existing methods primarily use asynchronous auto-regressive models or synchronous diffusion models to address this challenge. However, asynchronous auto-regressive models often suffer from inconsistencies between training and inference, leading to issues such as error accumulation, while synchronous diffusion models are limited by their reliance on rigid sequence length.  
To address these issues, we introduce Auto-Regressive Diffusion (AR-Diffusion), a novel model that combines the strengths of auto-regressive and diffusion models for flexible, asynchronous video generation. Specifically, our approach leverages diffusion to gradually corrupt video frames in both training and inference, reducing the discrepancy between these phases. Inspired by auto-regressive generation, we incorporate a non-decreasing constraint on the corruption timesteps of individual frames, ensuring that earlier frames remain clearer than subsequent ones. This setup, together with temporal causal attention, enables flexible generation of videos with varying lengths while preserving temporal coherence.
In addition, we design two specialized timestep schedulers: the FoPP scheduler for balanced timestep sampling during training, and the AD scheduler for flexible timestep differences during inference, supporting both synchronous and asynchronous generation.
Extensive experiments demonstrate the superiority of our proposed method, which achieves competitive and state-of-the-art results across four challenging benchmarks.
\renewcommand{\thefootnote}{}
\footnote{* denotes equal contributions.}
\renewcommand{\thefootnote}{\arabic{footnote}}
\footnotetext{This research was completed during the internship at bytedance.}
\footnote{\url{https://github.com/iva-mzsun/AR-Diffusion}.}
\footnote{Playable video samples: \url{https://iva-mzsun.github.io/AR-Diffusion}.}
\end{abstract} 
\section{Introduction}
\label{sec:intro}
Video generation aims to create sequences of frames that are both visually realistic and temporally consistent, ensuring that objects in the video appear vivid and their movements are smooth.
Current video generation methods \cite{VDM,lvdm,vidm,magicvideo,svd,latte,cogvideo,videogpt,magvit2,moso,videopoet,tats} can be divided into two categories: synchronous and asynchronous models, depending on how noise is applied during training.
Synchronous video generation models apply the same level of noise across all video frames during training, ensuring that each frame has the same signal-noise ratio.
Video diffusion models \cite{VDM,lvdm,vidm,magicvideo,svd,latte} are common examples of synchronous method, which apply a shared noise timestep to all frames.  These models have demonstrated promising results, but their reliance on equal noise levels and fixed-length sequences limits their ability to generate videos with varied lengths. 
To address these limitations, some recent studies have tried a ``chunked" approach, where multiple frames are generated simultaneously  based on a few preceding ones \cite{voleti2022mcvd,chen2023seine,luo2023videofusion,blattmann2023align}. While this approach helps to reduce computational complexity, it often suffers from temporal inconsistencies and motion discontinuities due to the limited temporal context.

\begin{figure}
    \centering
    \includegraphics[width=1.0\linewidth]{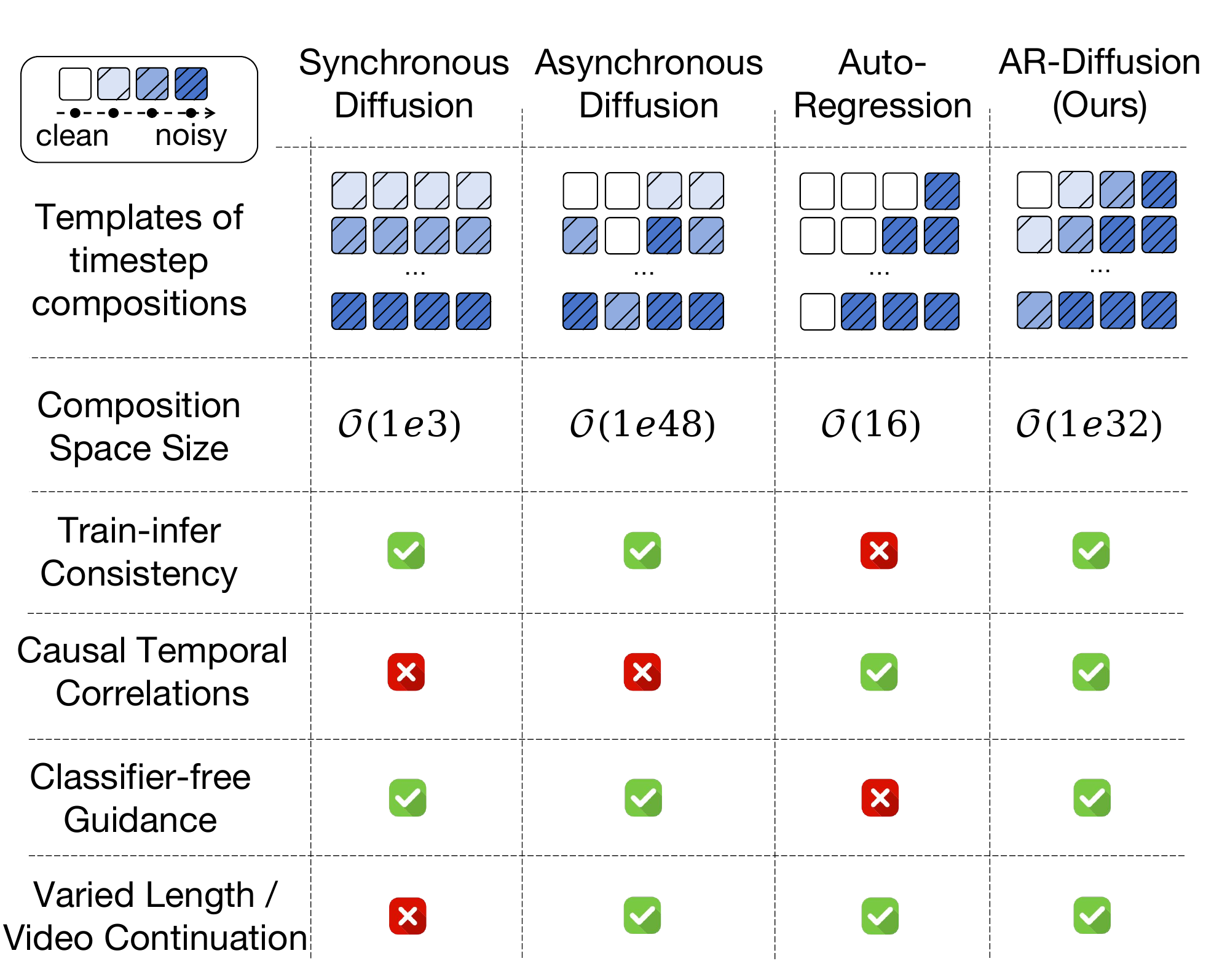}
    \vspace{-7mm}
    \caption{Different generative models employ different constraints on the timestep compositions and thus exhibit different properties.}
    \vspace{-6mm}
    \label{fig:intro}
\end{figure}

Asynchronous video generation models, in contrast, allow each frame to be corrupted by different noise levels, making them more adaptable to frames with varying complexities. 
For example, auto-regressive video generative models \cite{cogvideo,videogpt,moso,magvit2,videopoet,tats} generate each frame based on previous clear frames and add noise to future frames, which helps create variable-length videos. However, they often face inconsistencies between training and inference, causing issues like error accumulation \cite{glober}.
Asynchronous diffusion models \cite{diffusion_forcing,fvdm}, which apply independent noise to each frame, offer more flexibility but face challenges in training stability and convergence efficiency \cite{fvdm}.

To overcome these challenges, we propose Auto-Regressive Diffusion (AR-Diffusion). First, to make the generation process more efficient, we introduce an Auto-Regressive Video Auto-Encoder (AR-VAE), which compresses video frames into compact, continuous latent features. 
AR-VAE consists of a time-agnostic video encoder that extracts features without considering the temporal correlations, and a temporal causal video decoder that reconstructs the frames with temporal casual attention to maintain temporal consistency.
We then propose the AR-Diffusion framework, which combines the strengths of both auto-regressive and diffusion methods. 
By applying diffusion to gradually corrupt video frames during both training and inference, it reduces discrepancies between these phases, which helps prevent issues like error accumulation. Akin to auto-regressive models, AR-Diffusion employs temporal causal attention and allows earlier video frames to maintain clearer content compared to later ones through asynchronous frame-wise timesteps, thus supporting adaptive inference and facilitating the generation of videos of varied lengths. Unlike previous methods, AR-Diffusion applies a unique non-decreasing constraint on the noise levels: earlier frames remain clearer while later frames get more noise. This helps maintain a natural and coherent progression of generated content, resulting in smoother videos. As illustrated in \cref{fig:intro}, compared to existing asynchronous video diffusion models, our non-decreasing constraint can significantly reduce the search space of timestep compositions, thus improving training stability and speeding up convergence.

In addition, we introduce two specialized timestep schedulers. During training, we employ the Frame-oriented Probability Propagation (FoPP) scheduler to balance the sampling of timestep compositions and frame-specific timesteps, ensuring that the model can be well generalized to different inference settings.
During inference, we introduce the Adaptive-Difference (AD) scheduler, which allows the timestep difference between neighboring frames to vary adaptively. This flexibility supports both auto-regressive asynchronous and diffusion-based synchronous generation, enhancing adaptability during the inference process.  

We have conducted comprehensive experiments to validate the effectiveness and efficiency of our proposed method.
The results shows that our method achieves competitive and state-of-the-art (SOTA) performance on four challenging benchmarks, including FaceForensics \cite{faceforensics}, Sky-Timelapse \cite{xiong2018learning}, Taichi-HD \cite{siarohin2019first}, and UCF-101 \cite{ucf101}.
In particular, our AR-Diffusion surpasses the previous SOTA asynchronous video diffusion model \cite{fvdm} by 60.1\% FVD score on the UCF-101 dataset.
Furthermore, we explore the impact of varying timestep differences in video generation across multiple benchmarks, demonstrating that an appropriate timestep difference can significantly enhance the generation performance. 
Specifically, on the Taichi-HD dataset, an optimal timestep difference can bring a 14.6 and 5.4 improvement in the FVD score compared to synchronous and auto-regressive video generation inference settings, respectively.

Our contributions are summarized as follows: 
\begin{itemize}
    \item We introduce AR-VAE and propose a novel video generative model called AR-Diffusion, which combines the benefits of both asynchronous auto-regressive and synchronous diffusion models.
    \item We introduce a novel FoPP timestep scheduler during training, which balances uniform sampling of timestep compositions and frame-specific timesteps.
    \item We propose an AD video scheduler for inference, which allows the timestep difference between neighbor frames to be adaptive.
    \item Extensive experiments demonstrate the effectiveness and efficiency of our proposed method, validating the necessity of allowing asynchronous timesteps and the importance of finding an optimal timestep difference.
\end{itemize}

\section{Related Works}
\label{sec:relatedworks}

\subsection{Synchronous Video Generation}
Synchronous video generation models apply consistent noise or transformations across all video frames, maintaining uniform information entropy and temporal coherence throughout the sequence. Diffusion-based video generative models \cite{VDM,fdm,magicvideo, vidm, imagenvideo, makeavideo,videocrafter1,magicvideo2,svd,lavie,makepixeldance,latte,pvdm,lvdm,opensora,opensoraplan,cogvideox}, are commonly used in this category, as they employ an equal timestep scheduler during both training and inference to ensure consistency across frames. VDM \cite{VDM} was the first to apply diffusion models to video generation, which introduced a 3D U-Net architecture for video synthesis. Building upon this, models such as LVDM \cite{lvdm} and FDM \cite{fdm} adopted similar strategies, using identity noise timesteps across different video frames to ensure temporal consistency.
Imagen Video \cite{imagenvideo} and other subsequent works like Make-A-Video \cite{makeavideo}, Magic Video \cite{magicvideo} further improved upon these methods by incorporating spatio-temporally factorized architectures, enabling the generation of high-definition videos with more complex temporal dependencies. Recent models such as Lavie \cite{lavie}, PixelDance \cite{makepixeldance}, Open-Sora \cite{opensora} and Latte \cite{latte} have extended these techniques by utilizing larger training datasets and advanced model architectures to enhance generalization and visual fidelity.

In addition to diffusion-based methods, GAN-based approaches have also significantly contributed to synchronous video generation. A collection methods, such as MoCoGAN \cite{mocogan}, DIGAN \cite{digan}, StyleVideoGAN \cite{stylevideogan}, MoStGAN-V \cite{mostgan-v} and LongVideoGAN \cite{longvideogan}, achieve temporal consistency by synchronizing content representations across frames while allowing motion-specific variations. 
Although these methods have shown promising results in capturing temporal dynamics, GAN-based models face challenges such as mode collapse and training instability, which makes it difficult to maintain high-fidelity performance on challenging datasets.

To alleviate the limitations of the equal timestep scheduler, recent researchers have investigated chunked autoregressive generation based on diffusion models, which predict multiple frames in parallel based on a few preceding ones. Methods like MCVD \cite{voleti2022mcvd}, SEINE \cite{chen2023seine}, and VideoFusion \cite{luo2023videofusion} reduce computational complexity by generating frames in chunks. 
However, the temporal context in this approach is limited, thus resulting in inconsistent temporal dynamics.

\subsection{Asynchronous Video Generation}
Asynchronous video generation methods allow each video frame to be processed with distinct noise levels, offering greater adaptivity in handling varying information entropy across frames. 
These methods are particularly well-suited for generating videos with variable lengths, as they can be naturally extended to longer tokens or generate subsequent video frames in a first-in-first-out manner.
Auto-regressive video generative methods is a representative asynchronous approach, such as CogVideo \cite{cogvideo}, VideoGPT\cite{videogpt}, MOSO\cite{moso}, Magvit2 \cite{magvit2}, VideoPoet \cite{videopoet} and TATS \cite{tats}, which generate each frame sequentially by conditioning on the previously generated ones. This approach allows for flexible, variable-length generation but often introduces inconsistencies between training and inference, leading to issues such as error accumulation and degraded quality in long sequences \cite{glober}.

Asynchronous diffusion models offer a more flexible approach by applying independent noise timesteps to each video frame. For example, Diffusion Forcing \cite{diffusion_forcing} and FVDM \cite{fvdm} use frame-specific noise levels, which enhances sampling flexibility and enables better handling of temporal variability. However, these models still face challenges in achieving stable training \cite{fvdm} and are often outperformed by synchronous methods due to the expanded search space introduced by the independent noise timesteps for each frame.


\section{Methods}
\begin{figure*}
    \centering
    \includegraphics[width=1.0\linewidth]{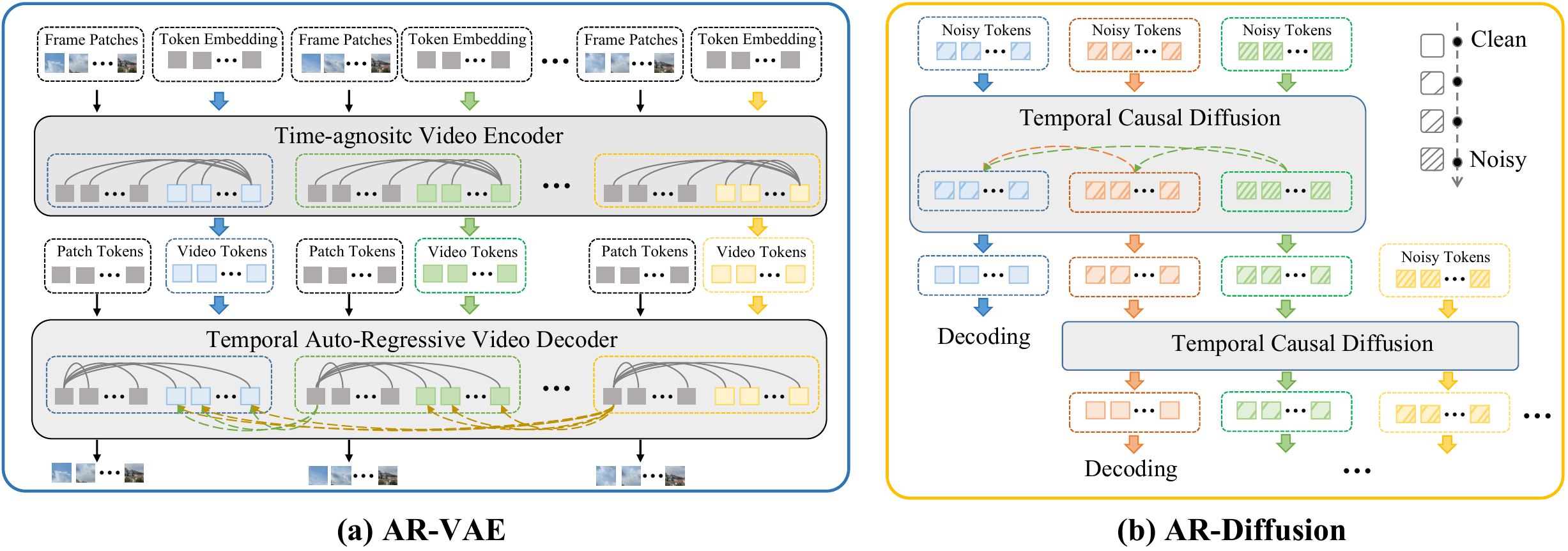}
    \vspace{-5mm}
    \caption{
    The overall framework of our proposed (a) AR-VAE and (b) AR-Diffusion.
    AR-VAE encodes videos into latent video tokens and AR-Diffusion models asynchronous video generation in the latent space.
    }
    \vspace{-5mm}
    \label{fig:framework}
\end{figure*}

In this section, we provide a detailed explanation of our proposed method. 
An overview of the entire framework is illustrated in Fig. \ref{fig:framework}.
We first describe our Auto-Regressive Video Auto-Encoder (AR-VAE) in Sec. \ref{sec:method_arvae}, and then introduce the AR-Diffusion video generative model in \ref{sec:method_ardiff}.
Next, we introduce the FoPP timestep scheduler in Sec. \ref{sec:method_fopp}, which is used during training to determine the composition of timesteps for corrupting input video frames.  Finally, we describe the AD timestep scheduler, which is used during inference to regulate the timestep compositions for video sampling, as detailed in Sec. \ref{sec:method_ad}.

We represent a video as $\textbf{x} \in R^{F \times H \times W \times C}$, where $F$ is the number of video frames, $H$ is the height, $W$ is the width, and $C$ is the number of channels.
Our AR-VAE encodes a video $\textbf{x}$ into video features $\textbf{z} \in R^{F \times L \times D}$, where $L$ and $D$ are the length and dimension of latent tokens, respectively.
$z_i$ represents the latent feature of the $i$-th video frame $x_i$.
We employ $T$ diffusion timesteps, and the noise timestep for the $i$-th video frame is denoted as $t_i$.
A possible timestep composition for all $F$ video frames can be represented as $\langle t_1,t_2,...t_{F} 
\rangle, 1 \leq t_i \leq T, 1 \leq i \leq F$.
Different diffusion models employ different constraints on these timestep compositions, which will be detailed in Sec. \ref{sec:method_ardiff}.

\subsection{AR-VAE}
\label{sec:method_arvae}

Our AR-VAE model is built upon on a Transformer-based 1-Dimensional Tokenizer (Titok) \cite{titok}, which encodes an image into $L$ discrete tokens. TiTok provides a more compact latent representation, resulting in substantially more efficient and effective encoding compared to conventional techniques. Therefore, our AR-VAE extends Titok's framework to video sequences.
As illustrated in \cref{fig:framework}(a), we first divide each video frame into frame patches. 
These frame patches are then concatenated with $L$ learnable video token embeddings into the Transformer-based video encoder.
Through cross-attention, these token embeddings can effectively capture the video content, thereby serving as visual tokens that represent each video frame.
In the original Titok, a vector-quantization module was used to quantize and discretize the image tokens. However, in AR-VAE, we remove this module and instead use the features output by the first normalization layer of the video decoder to represent videos, allowing continuous representation of video features.

During decoding, learnable patch tokens are used as placeholders for the patches of each video frame, and these tokens are shared across different frames. This mechanism ensures that each patch token can interact with others within the same frame to maintain consistency during reconstruction.
As shown in the gray solid lines in the video decoder in \cref{fig:framework}(a), each patch token is capable of interacting with other patch tokens and visual tokens of the current frame via full attention.
Here, we further modify the video decoder to be temporally causal, enhancing temporal coherence and facilitating potential future applications in image auto-encoding.
Specifically, patch tokens from subsequent video frames can refer to latent tokens from previous frames, which strengthens their temporal correlations and improves temporal consistency.
We represent such reference with dotted and directed lines in \cref{fig:framework}(a), indicating that the reference is unidirectional; In other words, video tokens in preceding frames cannot access patch tokens from subsequent video frames.
Finally, the video encoder and decoder are together optimized to reconstruct the frame patches.
To enhance the clarity of reconstructed video frames, we incorporate an additional adversarial training loss \cite{esser2021taming}.
Moreover, the dimensionality of the video features $D$ is reduced to facilitate the optimization of the follow-up AR-Diffusion by removing parameters associated with later dimensions.

\subsection{AR-Diffusion}
\label{sec:method_ardiff}
As illustrated in Fig. \ref{fig:framework}, AR-Diffusion employs a Transformer with temporal causal attention as its backbone. 
The AR-Diffusion model takes noisy tokens as input and is optimized to output clean tokens.
During training, clean video tokens are corrupted according to a sampled timestep composition, and $x_0$ prediction loss is used to optimize AR-Diffusion.
During inference, AR-Diffusion transforms a pure noise into a realistic video sample by iteratively performing the following steps:
1) Predict a clean sample based on the noisy input; 
2) Recorrupt the predicted clean sample based on the corresponding timestep composition;
3) Feed the corrupted result into the model for next-step prediction.
We discuss the necessity of temporal casual attention and $x_0$ prediction loss in the appendix.

\noindent \textbf{Diffusion Theory}
Following \cite{ldm}, each frame feature $z_i^0$ (i.e. $z_i$) is corrupted by $T$ steps during the forward diffusion process using the transition kernel:
\begin{gather}
q(z_i^{t_i}|z_i^{t_i-1}) = \mathcal{N}(z_i^{t_i}; \sqrt{1-\beta_{t_i}} z_i^{t_i-1}, \beta_{t_i} \textbf{I}) \label{eq:forward1} \\
q(z_i^{t_i}|z_i^0) = \mathcal{N}(z_i^{t_i}; \sqrt{\bar \alpha_{t_i}} z_i^0, (1 - \bar \alpha_{t_i}) \textbf{I}) \label{eq:forward2}
\end{gather}
where $\{ \beta_{t} \in (0, 1) \}_{{t}=1}^T$ is a set of hyper-parameters, $\alpha_{t} = 1 - \beta_{t}$ and $\bar \alpha_{t} = \prod_{i=1}^{t} \alpha_i$.
Based on Eq. (\ref{eq:forward2}), we can obtain the corrupted feature $z_i^{t_i}$ directly given the timestep $t_i$ during the training and inference phases as follows:
\begin{equation}
z_i^{t_i} = \sqrt{\bar \alpha_{t_i}} z_i^0 + (1 - \bar \alpha_{t_i}) \epsilon_{t_i}
\label{eq:forward_z}
\end{equation}
where $\epsilon_{t_i}$ is a noise feature sampled from an isotropic Gauss distribution $\mathcal{N}(0, \textbf{I})$.
The reverse diffusion process $q(z_i^{t_i-1}|z_i^{t_i}, z_i^0)$ has a traceable distribution:
\begin{equation}
    q(z_i^{t_i-1}|z_i^{t_i}, z_i^0) = \mathcal{N}(z_i^{t_i-1} | \tilde{\mu}_{t_i}(z_i^{t_i},z_i^0), \tilde{\beta}_{t_i} \textbf{I})
    \label{eq:reverse_diffusion}
\end{equation}
where $\tilde{\mu}_{t_i}(z_i^{t_i},z_i^0) = \frac{1}{\sqrt{\alpha_{t_i}}} (z_i^{t_i} - \frac{\beta_{t_i}}{\sqrt{1 - \bar \alpha_{t_i}}} \epsilon_{t_i} )$, $\epsilon_{t_i} \sim \mathcal{N}(0, \textbf{I})$, and $\tilde{\beta}_{t_i} = \frac{1 - \bar \alpha_{{t_i}-1}}{1 - \bar \alpha_{t_i}} \beta_{t_i}$.

During training, $x_0$ prediction optimizes the model to predict $z_i^0$ based on $z_i^{t_i}$.
During inference, we can obtain $z_i^{t_i-1}$ based on input $z_i^{t_i}$ and predicted $z_i^0$ using Eq. \ref{eq:reverse_diffusion}.


\noindent \textbf{Non-decreasing Timestep Constraint.} 
In synchronous diffusion models \cite{VDM}, a shared noise timestep is applied to all video frames during each training and inference step.
This constraint can be formulated as $ t_1 = t_2 = ... = t_F$.
In a typical scenario with $F=16$ and $T=1000$, the number of possible timestep compositions satisfying this constraint is $\mathcal{O}(1e3)$, which significantly limits the flexibility of the model.
Conversely, recent asynchronous diffusion models \cite{diffusion_forcing} allow each video frame to have an independently sampled noise timestep, expanding the number of possible timestep compositions to $\mathcal{O}(1e48)$.
However, this vast search space introduces redundancy, as many timestep combinations are never utilized during inference, leading to training instability \cite{fvdm}, as discussed in the appendix.

To overcome these challenges, we propose a non-decreasing timestep constraint, which can be formulated as $t_1 \leq t_2 \leq ... \leq t_F$.
Intuitively, this constraint ensures that earlier video frames remains as clear or clearer than subsequent frames.
Under this constraint,  the number of valid timestep compositions in the above-mentioned typical scenario becomes $\mathcal{O}(1e32)$, providing greater diversity than the equal constraint while being significantly more stable than the independent sampling approach.

\begin{figure}
    \centering
    \includegraphics[width=1.0\linewidth]{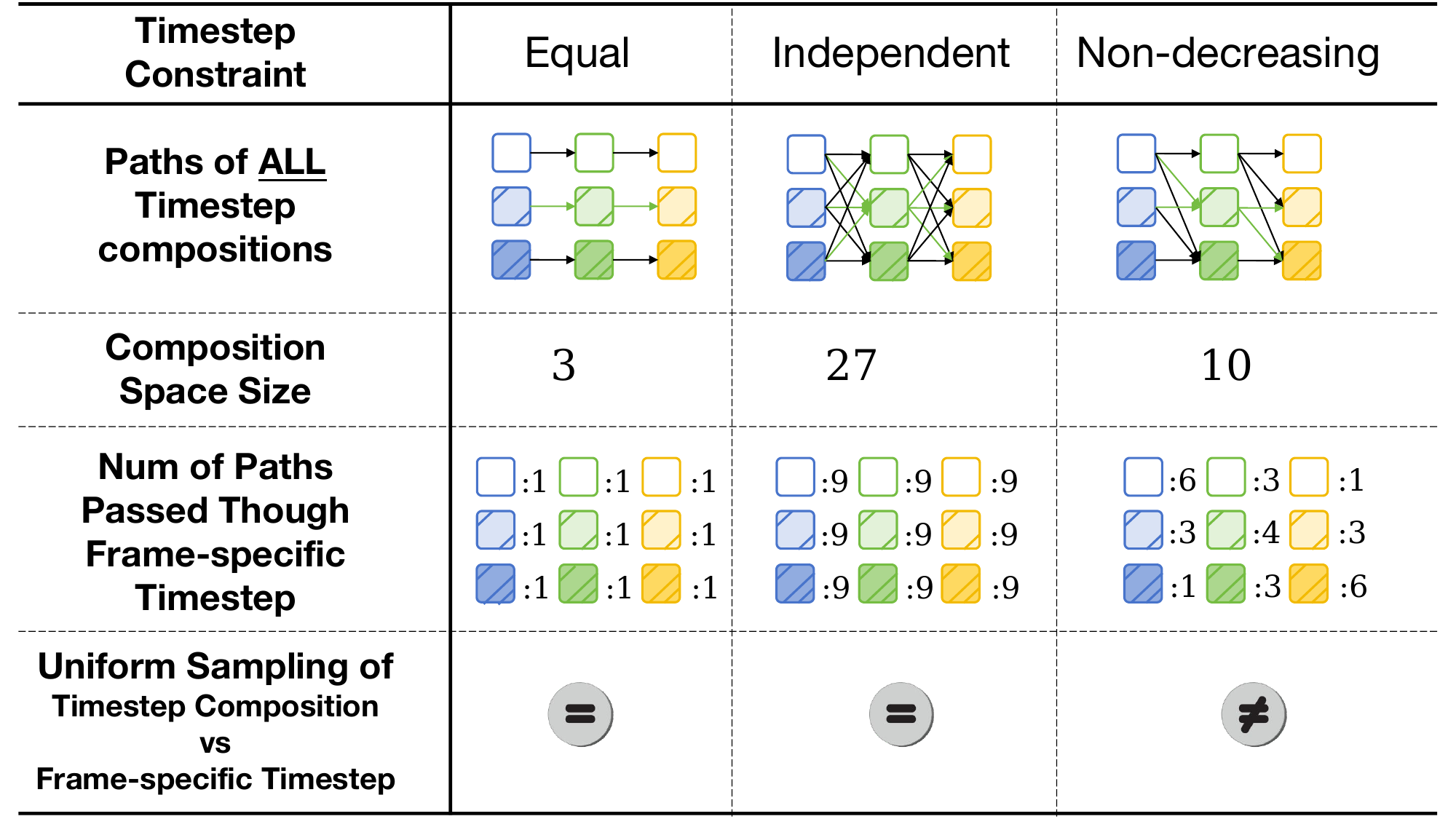}
    \vspace{-7mm}
    \caption{Settings and properties of different timestep constraints when the number of frames and timesteps is set to be $F=3$ and $T=3$, respectively.}
    \label{fig:timestep_constraint}
    \vspace{-7mm}
\end{figure}

\subsection{FoPP Timestep Scheduler}
\label{sec:method_fopp}
A significant challenge in training our AR-Diffusion is the demand for an appropriate timestep scheduler.
During training, the timestep scheduler determines the specific timestep composition utilized to corrupt video frames.
It is crucial since it determines how generative models can be generalized to different inference settings.
Existing synchronous and asynchronous diffusion models employ timestep schedulers that either equally or independently sample timesteps for video frames.
Specifically, synchronous diffusion models adopt an equal timestep scheduler, which first selects a random $t$ from the range of $1$ to $T$ and then applies this $t$ as the noise timestep across all video frames. 
Conversely, asynchronous diffusion models implement a frame-wise independent timestep scheduler. 
This involves independently sampling a noise timestep $t_i \sim \mathcal{U}(1,T)$ for each video frame sequentially, where $i$ ranges from $1$ to $F$.
Both equal and independent timestep schedulers naturally enable both uniformly sampling of timestep compositions and uniformly sampling of frame-specific timesteps, as illustrated in \cref{fig:timestep_constraint}.
However, the imposition of the non-decreasing constraint brings conflict between these two sampling uniformity in our AR-Diffusion, as it leads to varying frequencies of frame-specific timesteps visited by all possible timestep compositions.
To this end, a novel timestep scheduler is required that achieves a balance between these two sampling uniformity when training our AR-Diffusion.

Given previous timestep schedulers, we intuitively design our timestep scheduler by uniformly sampling the noise timestep $t_1 \sim \mathcal{U}(1,T)$ for the first video frame, followed by $t_2 \sim \mathcal{U}(t_1, T)$ for the second frame, and continuing in this manner for subsequent frames.
However, this approach results in significantly unbalanced probabilities for different timestep compositions. 
For instance, if the timestep for the first frame is sampled at $T$, which occurs with a probability of $\frac{1}{1000}$, then all subsequent frame timesteps must also be $T$, due to our non-decreasing timestep constraint. 
Consequently, the composition $\langle t_1=T, t_2=T, ..., t_F=T \rangle$ is sampled with a probability of $\frac{1}{1000}$. 
Given that there are approximately $\mathcal{O}(1e32)$ possible timestep compositions in the typical setting, this probability is excessively high, leading to severe bias in the training process.

\begin{algorithm}[t]
\label{alg:fopp} \small
\caption{Algorithm of the FoPP timestep scheduler}
\begin{algorithmic}[1]
\Require $T$ (Total timesteps), $F$ (Total frames)
\Ensure $\langle t_1, t_2, ..., t_F \rangle$
\Statex Initializing matrixes $d^s, d^e \in N^{F \textbf{x} T}$
\Statex Randomly select $f \sim \mathcal{U}(1,F)$ and $t_f \sim \mathcal{U}(1,T)$ \\
// Sample timesteps for previous video frames
\For{$i = f-1$  to $1$}
    \Statex $P^e = d^e[i, 1:t_{i+1}+1] / sum(d^e[i, 1:t_{i+1}+1])$
    \Statex Sample $t_f$ from $\{1,...,t_{i+1}\}$ based on probability $P^e$
\EndFor \\
// Sample timesteps for subsequent video frames
\For{$i = f+1$ to $F$}
    \Statex $P^s = d^s[i, t_{i-1}:T+1] / sum(d^s[i, t_{i-1}:T+1])$
    \Statex Sample $t_f$ from $\{ t_{i-1},...,T \}$ based on $P^s$
\EndFor
\end{algorithmic}
\end{algorithm}

To tackle this issue, we introduce the Frame-oriented Probability Propagation (FoPP) timestep scheduler. 
Firstly, we uniformly sample a frame index $f \sim \mathcal{U}(1, F)$ and a corresponding timestep $t \sim \mathcal{U}(1, T)$, ensuring a uniform distribution of timesteps across all video frames. 

Then, we utilize the methodology of dynamic programming to calculate and propagate the probability of each timestep for previous and subsequent video frames that can be visited by all possible timestep compositions.
In this way, we uniformly sample a timestep composition conditioned on $t_f=t$.
In particular, we first use $d^s_{i,j}, 1 \leq i \leq F, 1 \leq j \leq T$ to represent the total number of timestep compositions that satisfies $\langle t_i=j, t_{i+1}, ..., t_F \rangle$ and the non-decreasing constraint. 
Here, $\langle t_i=j, t_{i+1}, ..., t_F \rangle$ means that each timestep composition is starting with the $i$-th video frame at timestep $t_i=j$.
Then, we calculate $d^s_{i,j}$ with $i$ from $F$ to 1 and $j$ from $T$ to 1 using dynamic programming, where $d_{*,T}=1$, $d_{F,*}=1$, and the transition equation being $d_{i,j}=d_{i,j-1}+d_{i-1,j}$. 
Then the visit probability of the timestep $k\in[K,T]$ of the subsequent video frame $i$, given $i>f$ and $t_{i-1}=K$, is $\frac{d^s_{i,k}}{\sum_{j=K}^T d^s_{i,j}}$.
Similarly, we calculate $d^e_{i,j}$, which represents the total number of timestep compositions that satisfies $\langle t_1, t_2, ..., t_i=j \rangle$ and the non-decreasing constraint. 
Then the visit probability of the timestep $k\in[1,K]$ of the previous video frame $i$, given $i<f$ and $t_{i+1}=K$, is $\frac{d^e_{i,k}}{\sum_{j=1}^K d^e_{i,j}}$.

Finally, the timesteps of previous or subsequent video frames are sampled one by one based on corresponding probabilities.
The methodology is detailed in Algorithm 1. 

\subsection{AD Timestep Scheduler}
\label{sec:method_ad}
During inference, a timestep scheduler is utilized to regulate a sequence of timestep compositions that starts from $\langle t_1=T,t2=T,...,t_F=T \rangle$ to $\langle t_1=0,t2=0,...,t_F=0 \rangle$.
In other words, it determines which timestep composition is used in each sampling step.
Here, we introduce an AD timestep scheduler, which supports adaptive video generation and accommodates both asynchronous auto-regressive and synchronous generation.
Specifically, the AD timestep scheduler makes the timestep difference between neighboring video frames as an adaptive variable $s$. 
For timesteps $t_i$ and $t_{i-1}$ of consecutive video frames, we ensure that the following condition holds:
\begin{equation}
  t_i = 
  \begin{cases}
	t_i + 1, &\text{if $i=1$ or $t_{i-1} = 0$,} \\
    min(t_{i-1}+s, T), &\text{if $t_{i-1} > 0$} 
  \end{cases}
\end{equation}
Specifically, when the previous video frame is none or clean, the current video frame focuses on the denoising of itself; 
Otherwise, it keeps denoising with a timestep difference of $s$ than its previous video frame.
Notably, the synchronous diffusion model and the auto-regressive generative model represent two distinct cases within our variable difference timestep scheduler, corresponding to $s=0$ and $s=T$, respectively. 
Intuitively, a smaller $s$ results in neighboring video frames having more similar content, whereas a larger $s$ introduces greater content variability. 
Importantly, we apply a mask to those frame latents for which the noise timesteps remain unchanged.

\begin{table*}[t] \small
    \centering
    \caption{
    Quantitative comparison on four challenging datasets: Taichi-HD \cite{siarohin2019first}, Sky-Timelapse \cite{xiong2018learning}, FaceForensics \cite{faceforensics}, and UCF-101 \cite{ucf101}.
    }
    \vspace{-3mm}
    \label{tab:quantitative_comparison}
    \scalebox{0.8}{
        \begin{tabularx}{1.2\linewidth}{ 
            p{0.2\linewidth} | 
            X<{\centering} | X<{\centering} | 
            X<{\centering} | X<{\centering} | 
            X<{\centering} | X<{\centering} | 
            X<{\centering} | X<{\centering}}
        \toprule
            & \multicolumn{2}{c|}{Taichi-HD \cite{siarohin2019first}} 
            & \multicolumn{2}{c|}{Sky-Timelapse \cite{xiong2018learning}}
            & \multicolumn{2}{c|}{FaceForensics \cite{faceforensics}} 
            & \multicolumn{2}{c}{UCF-101 \cite{ucf101}} \\
        \midrule
        & FVD$_{16}$ & FVD$_{128}$ & FVD$_{16}$ & FVD$_{128}$ & FVD$_{16}$ & FVD$_{128}$ & FVD$_{16}$ & FVD$_{128}$ \\
        \midrule
        \rowcolor{yellow!10} \multicolumn{9}{c}{Generative Adversarial Models}      \\
        MoCoGAN \cite{mocogan}          &-      &-      &206.6  &575.9  &124.7  &257.3  &2886.9 &3679.0  \\ 
        + StyleGAN2 backbone            &-      &-      & 85.9  &272.8  &55.6   &309.3  &1821.4 &2311.3  \\
        MoCoGAN-HD \cite{mocogan-hd}    &-      &-      &164.1  &878.1  &111.8  &653.0  &1729.6 &2606.5  \\
        DIGAN \cite{digan}              &128.1  &748.0  &83.11  &196.7  &62.5   &1824.7 &1630.2 &2293.7  \\
        StyleGAN-V
        \cite{skorokhodov2022stylegan}  &143.5  &691.1  &79.5   &197.0  &47.4   &89.3   &1431.0 &1773.4  \\
        MoStGAN-V \cite{mostgan-v}      &-      &-      &65.3   &162.4  &39.7   & \textbf{72.6}  &-      &-       \\ 
        \midrule
        \rowcolor{green!10} \multicolumn{9}{c}{Auto-regressive Generative Models}      \\
        VideoGPT \cite{videogpt}        &-      &-      &222.7  &-      &185.9  &-      &2880.6 &-       \\
        TATS \cite{tats}                &94.6   &-      &132.6  &-      &-      &-      &420    &-       \\
        \midrule
        \rowcolor{blue!10} \multicolumn{9}{c}{Synchronous Diffusion Generative Models}      \\
        Latte \cite{latte}              &159.6  &-      &59.8   &-      &\textbf{34.0}   & -     &478.0  &-       \\
        PVDM \cite{pvdm}                &540.2  &-      &55.4   &\textbf{125.2}  &355.9  & -     &343.6  &648.4 \\ 
        LVDM \cite{lvdm}                &99.0   &-      &95.2   &-      &-      &-      &372.9  &-      \\ 
        VIDM \cite{vidm}                &121.9  &563.6  &57.4   &140.9  &-      &-      &294.7  &1531.9 \\ 
        \midrule
        \rowcolor{orange!10} \multicolumn{9}{c}{Asynchronous Diffusion Generative Models}      \\
        FVDM \cite{fvdm}                &194.6  &-      &106.1  &-      &55.0   &-      &468.2  &-         \\
        Diffusion Forcing 
        \cite{diffusion_forcing}        &202.0  &738.5  &251.9  &895.3  &99.5   &555.2  &274.5  &836.3   \\
        AR-Diffusion (ours)             &\textbf{66.3}   & \textbf{376.3} &\textbf{40.8}   &175.5  &71.9   &265.7  &\textbf{186.6}  & \textbf{572.3}       \\
        \bottomrule
        \end{tabularx}
    }
\end{table*}

\begin{table*}[t] \small
    \centering
    \caption{
    Quantitative results with different values of timestep $s$.
    We utilize a DDIM sampler with 50 sampling steps in total.
    }
    \vspace{-3mm}
    \label{tab:difference_timesteps}
    \scalebox{0.8}{
        \begin{tabularx}{1.2\linewidth}{ 
            p{0.05\linewidth}<{\centering} | 
            p{0.07\linewidth}<{\centering} p{0.07\linewidth}<{\centering} X<{\centering} | 
            p{0.07\linewidth}<{\centering} p{0.07\linewidth}<{\centering} X<{\centering} | 
            p{0.07\linewidth}<{\centering} p{0.07\linewidth}<{\centering} X<{\centering} | 
            p{0.07\linewidth}<{\centering} p{0.07\linewidth}<{\centering} X<{\centering} | 
            p{0.08\linewidth}<{\centering} }
        \toprule
            \multirow{2}{*}{$s$} & \multicolumn{3}{c|}{FaceForensics \cite{faceforensics}} & \multicolumn{3}{c|}{Sky-Timelapse \cite{xiong2018learning}} 
            & \multicolumn{3}{c|}{Taichi-HD \cite{siarohin2019first}} & \multicolumn{3}{c|}{UCF-101 \cite{ucf101}} 
            & \multirow{2}{*}{\makecell{Inference \\ Time (s)}} \\
        \cline{2-13}
            &FID-img&FID-vid&FVD &FID-img&FID-vid&FVD &FID-img&FID-vid&FVD &FID-img&FID-vid&FVD & \\
        \midrule
        \rowcolor{gray!30} \multicolumn{14}{c}{16-frame Video Generation}      \\
        0   & 14.0  & 6.9   & \textbf{71.9}  & 10.0  & \textbf{9.2}   & \textbf{40.8}  & 13.8  & 9.2   & 80.9  & 30.3  & 17.6  & 194.4 & 2.4  \\
        5   & 14.0  & 6.2   & 78.1  & 11.1  & 11.6  & 55.2  & 13.0  & 5.9   & \textbf{66.3}  & 30.0  & 17.7  & 194.0 & 5.2 \\
        10  & \textbf{13.6}  & \textbf{6.1}   & 84.4  & 10.3  & 11.2  & 57.6  & 12.4  & \textbf{5.8}   & 70.9  & 30.1  & 18.8  & 212.2 & 7.9 \\
        15  & 14.3  & 6.6   & 83.5  & 9.4   & 11.4  & 55.6  & \textbf{12.2}  & \textbf{5.8}   & 69.4  & 30.0  & \textbf{16.3}  & \textbf{186.6} & 10.8 \\
        20  & 14.8  & 6.3   & 83.3  & \textbf{9.2}   & 10.9  & 56.3  & 12.7  & 6.0   & 67.0  & 31.0  & 17.4  & 201.1 & 13.6 \\
        25  & 14.1  & \textbf{6.1}   & 79.0  & 9.7   & 10.5  & 48.4  & 12.9  & 6.5   & 75.1  & 29.6  & 17.3  & 191.6 & 16.4 \\
        50  & 14.2  & \textbf{6.1}   & 82.8  & 10.1  & 10.7  & 50.6  & 13.1  & 5.9   & 71.7  & \textbf{29.5}  & 17.1  & 192.6 & 30.5  \\
        \rowcolor{gray!30} \multicolumn{14}{c}{128-frame Video Generation}      \\
        5   & \textbf{14.7}  & 9.5   & \textbf{265.7} & 12.2  & 25.2  & 185.1 & 8.9   & \textbf{10.8}  & \textbf{376.3} & 32.5  & 24.4  & 592.7 & 42.1  \\
        10  & 15.2  &\textbf{ 8.9}   & 278.1 & \textbf{12.1}  & 23.9  & 182.6 & \textbf{8.8}   & 12.2  & 401.9 & \textbf{31.5}  & 24.9  & 605.3 & 78.0 \\
        25  & 15.4  & 9.3   & 348.6 & 12.2  & \textbf{22.8}  & \textbf{175.5} & \textbf{8.8}   & 12.3  & 402.5 & 31.8  & \textbf{23.3}  & \textbf{572.3} & 184.8 \\
        \bottomrule
        \end{tabularx}
    }
\end{table*}

\section{Experiments}
\begin{figure*}
    \centering
    \includegraphics[width=1.0\linewidth]{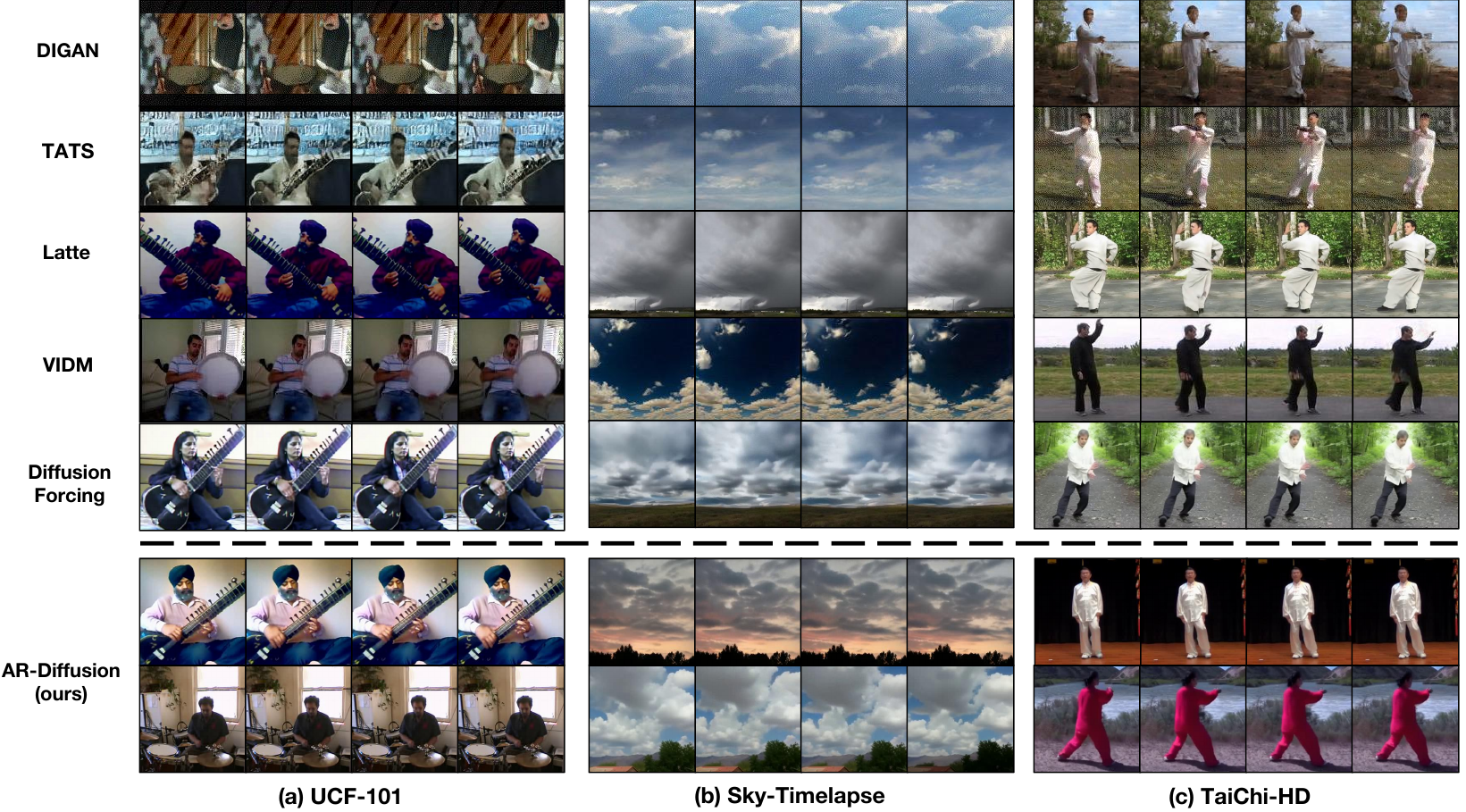}
    \vspace{-5mm}
    \caption{Qualitative comparison of existing video generative methods and our AR-Diffusion. }
    \vspace{-4mm}
    \label{fig:qualitative}
\end{figure*}

\subsection{Setups}

\noindent \textbf{Datasets.}
Following previous video generation methods \cite{moso,glober}, we conduct our experiments on four challenging datasets: FaceForensics \cite{faceforensics}, Sky-Timelapse \cite{xiong2018learning}, Taichi-HD \cite{siarohin2019first}, and UCF-101 \cite{ucf101}.

\noindent \textbf{Evaluation metrics.}
For quantitative evaluation, we use FID-img, FID-vid, and FVD \cite{unterthiner2018towards} to measure the realism of generated videos. 
For all metrics, the lower the score, the better the effect.
We consider two different clip lengths (16 and 128) for FVD, where we denote $FVD_{16}$ and $FVD_{128}$ as the FVD score measured on video clips of lengths 16 and 128, respectively.

\noindent \textbf{Implementation Details.}
Our experiments were conducted on videos with a resolution of $256^2$ and on 8 NVIDIA A800-40G GPUs.
For the AR-VAE training, we configured the length and dimension of the video latent features at $L=32$ and $D=4$, respectively. 
We initialized our AR-VAE using the publicly available checkpoint from Titok \cite{titok}. 
Initially, each AR-VAE was trained for 50,000 steps on individual video frames using typical reconstruction loss \cite{dalle}.
Subsequently, the adversarial loss was incorporated for an additional 450,000 training steps to further refine the model. 
Finally, we integrated temporal causal attention into our video decoder within the AR-VAE and conducted optimization across all parameters over an additional 100,000 steps.
The batch size is 16 per GPU and the learning rate is 1e-4.
We developed our AR-Diffusion using a Transformer-based diffusion model, specifically DiT \cite{dit}. 
Initially, video latent features are rescaled by a factor of 0.5 before input into AR-Diffusion. 
The training configuration includes a batch size of 16 per GPU, with gradient accumulation over 8 steps.
The learning rate is set at 2e-4 for the initial 100,000 training steps, followed by a fine-tuning phase for an additional 50,000 steps at a reduced rate of 1e-5. 
Techniques such as exponential moving average over all paramerters, gradient clipping, and noise clipping are also utilized.

\subsection{Quantitaive Comparison}
\label{sec:timestep_difference}
We compare our proposed AR-Diffusion with other methods on four challenging datasets, including FaceForensics \cite{faceforensics}, Sky-Timelapse \cite{xiong2018learning}, Taichi-HD \cite{siarohin2019first} and UCF-101 \cite{ucf101}. As shown in Table \ref{tab:quantitative_comparison}, our AR-Diffusion model achieves lower $FVD_{16}$ and $FVD_{128}$ scores compared to other generative methods, which indicates better video quality and temporal consistency. For example, on the FaceForensics dataset, AR-Diffusion achieves an $FVD_{16}$ of 71.9 and an $FVD_{128}$ of 265.7, respectively, outperforming most of the other models, including PVDM and Diffusion Forcing. On the Sky-Timelapse dataset, AR-Diffusion achieves an $FVD_{16}$ of 40.8, which is lower than all the other methods, showcasing its superior performance in generating high-quality temporal sequences. On the Taichi-HD dataset, AR-Diffusion achieves an $FVD_{16}$ of 66.3 and an $FVD_{128}$ of 376.3, outperforming all the other models including synchronous and asynchronous diffusion-based models, as well as generative adversarial models. 
On the more complex UCF-101 dataset, AR-Diffusion achieves an $FVD_{16}$ of 186.6 and an $FVD_{128}$ score of 572.3, which significantly outperforms all the other models. These results clearly demonstrate the superiority of AR-Diffusion across diverse datasets, establishing a new benchmark in video generation quality and consistency.

Table \ref{tab:difference_timesteps} presents quantitative results for AR-Diffusion with different settings of the timestep difference $s$, which controls the generation process and provides a trade-off between generation performance and efficiency.
Notably, $s=0$ corresponds to synchronous diffusion generation, while $s=50$ represents auto-regressive generation. The table reports performance on three metrics: FID-img, FID-vid, and FVD for both 16-frame and 128-frame video generation. 
For 16-frame video generation, AR-Diffusion generally achieves competitive FVD scores with $s$ values between 0 and 15, depending on the dataset. For example, on the FaceForensics and Sky-Timelapse datasets, the best FVD scores are achieved when $s=0$, indicating that synchronous generation provides optimal quality in this context. On Taichi-HD, $s=10$ provides the lowest FVD scores 66.3, suggesting that a moderate combination of synchronous and auto-regressive steps yields improved performance for this dataset. On UCF-101, the lowest FVD score of 186.6 is obtained with $s=15$, showing the effectiveness of moderate timestep values in balancing temporal coherence and quality for complex datasets.  As $s$ increases beyond $15$, FVD scores tend to increase, which suggests a reduction in the temporal quality of generated videos. In addition, inference time also increases with larger $s$ values, indicating a trade-off between efficiency and the use of auto-regressive steps. 
For 128-frame video generation, $s=5$ achieves relatively lower FVD scores, demonstrating that a smaller $s$ value is beneficial for generating longer sequences while maintaining visual quality. 
Overall, the analysis indicates that a moderate value of $s$ (e.g., $s=10$ for 16-frame generation and $s=5$ for 128-frame generation) often strikes an effective balance between quality and inference efficiency across different datasets.

\subsection{Qualitative Comparison}

We qualitatively compare our method with prior works as in Figure \ref{fig:qualitative}.
Samples of prior methods are obtained from \cite{vidm,luo2023videofusion} or generated with their released codes and parameters \cite{latte}.
On UCF-101, the GAN-based method DIGAN tends to produce video samples that lack distinctiveness. 
In contrast, TATS, which leverages a Transformer architecture and an interpolation strategy, creates more recognizable video samples. 
VIDM results in overly simplistic object motions.
The asynchronous diffusion model Diffusion Forcing \cite{diffusion_forcing} yields videos with minimal motion, while our AR-Diffusion method produces samples that showcase both clear appearances and notable movements.
On the Sky Time-lapse dataset, samples from DIGAN and TATS exhibit minimal motion and simplistic object representations. Conversely, Latte and VIDM enhance the detail and clarity of object boundaries, though their motion effects are still limited. 
In contrast, our AR-Diffusion method creates video samples characterized by dynamic movements and significantly enhanced visual details.
Similarly, for the TaiChi-HD dataset, video samples from DIGAN and TATS show noticeable distortions in human appearances. 
While Latte, VIDM, and Diffusion Forcing \cite{diffusion_forcing} improve the depiction of human figures, their movements remain minimal. 
In contrast, samples produced by our AR-Diffusion method display significant movements and clearly defined object appearances.

\subsection{Ablation Study}
\begin{table}[t] \small
    \centering
    \caption{
    Ablation study on our AR-Diffusion.
    }
    \vspace{-3mm}
    \label{tab:ablation}
    \scalebox{0.8}{
        \begin{tabularx}{1.2\linewidth}{ 
            p{0.50\linewidth} | X<{\centering} | X<{\centering} | X<{\centering} }
        \toprule
            &FID &FVD-img &FVD   \\
        \midrule
        AR-Diffusion                     & 12.2  & 13.4  & 62.8   \\
        -FoPP Timestep Scheduler    & 11.0  & 16.8  & 101.0  \\
        -Improved VAE               & 13.1  & 29.6  & 148.3  \\
        -Temporal Causal Attention  & 15.9  & 50.2  & 209.8  \\
        -x0 Prediction Loss         & 27.9  & 58.0  & 257.6  \\
        -Non-decreasing Constraint  & 32.2  & 87.9  & 272.5  \\
        \bottomrule
        \end{tabularx}
    }
\vspace{-3mm}
\end{table}
To systematically evaluate the contributions of different model components in our proposed AR-Diffusion model, we design various ablation models as shown in Table \ref{tab:ablation}. 
Each ablation experiment is conducted on the Sky-timelapse dataset using 8 A800 GPUs without the fine-tuning stage.
The baseline AR-Diffusion model achieves an FID of 12.2, an FVD-img of 13.4 and an FVD of 62.8, indicating high video quality and temporal consistency. 
Removing the FoPP timestep scheduler leads to an increase in FVD to 101.0, demonstrating the necessity of appropriately sampling timestep composition during training. 
Our AR-VAE is improved by reducing the latent dimension from 12 to 4, employing temporal causal attention, and incorporating the adversarial training loss.
Removing these improvements, the FVD score further increases to 148.3, illustrating their significant impact on generating high-quality video content. 
Eliminating temporal causal attention in the AR-Diffusion causes a substantial performance drop, with FID increasing to 15.9 and FVD rising to 209.8, underscoring its importance in maintaining temporal dependencies between frames. 
Replacing $x_0$ prediction with $\epsilon$ prediction has a particularly drastic effect, with FID increasing to 27.9 and FVD reaching 257.6, suggesting that this loss function is crucial for stable and consistent video generation. Finally, the absence of the non-decreasing constraint leads to the worst performance, with FID-img and FID-vid increasing to 32.2 and 87.9, respectively, and FVD reaching 272.5, indicating that this constraint is essential for maintaining quality during diffusion steps. Overall, the ablation study demonstrates that each component significantly contributes to the performance of AR-Diffusion, particularly the temporal causal attention, improved VAE, and $x_0$ prediction loss.

\section{Conclusion}
In this paper, we present a novel auto-regressive video diffusion model, which is called AR-Diffusion.
Targeting asynchronous video generation, AR-Diffusion introduces a novel non-decreasing timestep constraint, which significantly reduces the search space and thus stabilizes the training procedure.
During training, the FoPP timestep scheduler is used to optimize AR-Diffusion with a balanced sampling of timestep compositions.
During inference, the AD timestep scheduler is employed to regulate a sequence of timestep compositions for AR-Diffusion to produce a video sample.
In conclusion, AR-Diffusion combines the advantages of both auto-regressive generative models and synchronous video diffusion models, obtaining competitive or SOTA results on four challenging benchmarks.


{
    \small
    \bibliographystyle{ieeenat_fullname}
    \bibliography{main}
}

\appendix
\newpage


\section{Discussions}
\noindent \textbf{Discussion: Why temporal causal attention over bidirectional temporal attention?}
We incorporate the temporal causal attention mechanism for three main reasons: \textbf{1) Reducing noise interference}: Subsequent video frames often contain more noise and less information than preceding frames.
When preceding frames engage in cross-attention with subsequent frames, their information can be corrupted by the noise contained in subsequent frames.
\textbf{2) Auto-Regressive flexibility}: Temporal causal attention enables models to behave similarly to an auto-regressive model, which is well-suited for generating videos of variable lengths.
\textbf{3) Potential for image integration}: Temporal causal attention provides the capability to integrate image data as an initial video frame in future training, allowing for exclusive optimization of the starting frame.

\noindent \textbf{Discussion: Why use $x_0$ prediction instead of $\epsilon$ prediction or $v$ prediction?}
In many diffusion models, $\epsilon$ prediction \cite{ho2020denoising} and $v$ prediction \cite{salimans2022progressive} losses are more frequently employed. 
However, $x_0$ prediction is crucial for AR-Diffusion to effectively learn temporal correlations.
In synchronous diffusion models, video frames are uniformly corrupted with equal timesteps, preserving most temporal correlations, allowing the model to directly learn temporal relationships from the input. In contrast, asynchronous diffusion disrupts these correlations due to varying levels of corruption across frames, making it challenging for AR-Diffusion to learn temporal dependencies from the inputs alone. The use of $x_0$ prediction forces the model to generate outputs that maintain strong temporal correlations across frames, leading to improved video consistency.

\section{Limitation}
Despite the promising results achieved by our proposed AR-Diffusion, there are several limitations that need to be addressed in future work.
The primary limitation is that, while our model leverages video data for training, there is potential to further enhance its performance by incorporating image data. 
Images, being more readily available and diverse, can provide additional training signals that help improve the visual quality and diversity of generated frames. 
Integrating image data into the training process could also help in scenarios where video data is scarce or difficult to obtain. 
Future research should explore methods to effectively combine image and video data during training to further boost the performance of AR-Diffusion.

\section{Training Stability}
\begin{figure}
    \begin{minipage}[h]{1\linewidth}
        \centering
        \includegraphics[width=0.8\linewidth]{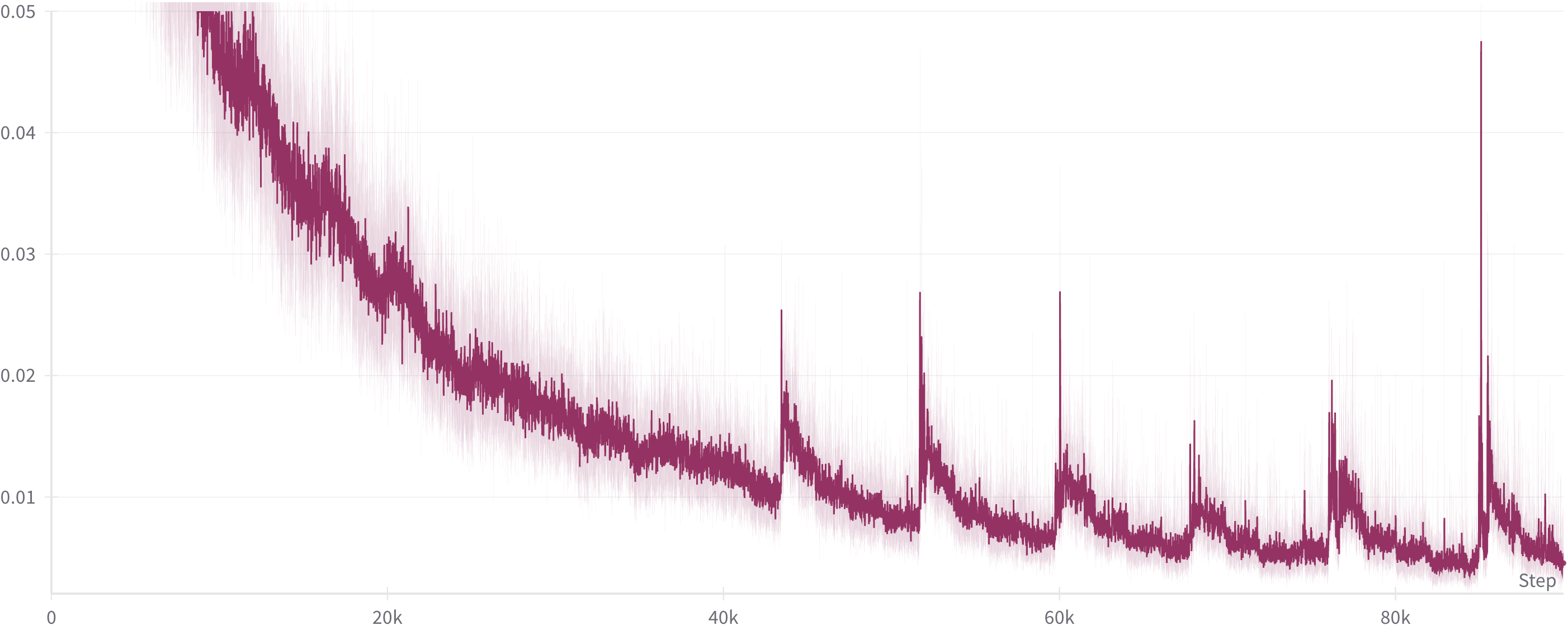}
        \figcaption{Loss curve of Diffusion Forcing \cite{diffusion_forcing} on UCF-101.}
        \label{fig:loss_curve_forcing}
    \end{minipage}
    \begin{minipage}[h]{1\linewidth}
        \centering
        \includegraphics[width=0.8\linewidth]{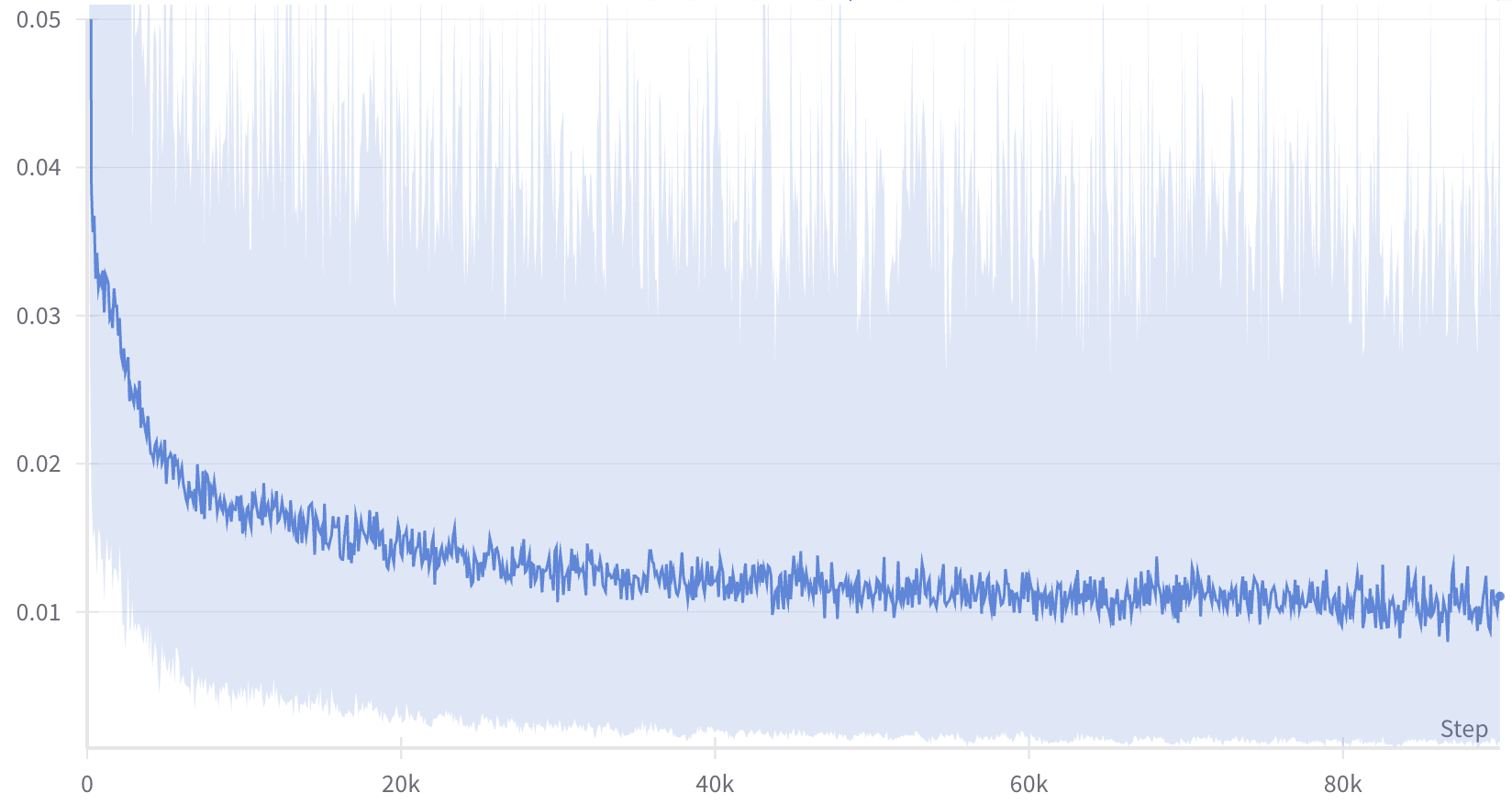}
        \figcaption{Loss curve of AR-Diffusion (ours) on UCF-101.}
        \label{fig:loss_curve_ardiff}
    \end{minipage}
\end{figure}

Training stability is a critical aspect of machine learning model performance. 
Stable training processes ensure consistent learning and convergence to optimal solutions. 
This paper explores the training stability of Diffusion Forcing \cite{diffusion_forcing} and our AR-Diffusion by analyzing their training loss curves.
In particular, we visualize their training loss curves in \cref{fig:loss_curve_forcing} and \cref{fig:loss_curve_ardiff}, respectively.
The training loss curves are plotted over a series of training steps. 
For Diffusion Forcing, the training loss decreases with noticeable fluctuations. 
These fluctuations suggest some instability in the training process.
Significant spikes in the loss are observed around 40k and 80k steps, indicating moments of instability. 
Despite these spikes, the overall trend shows a decrease in loss, suggesting ongoing learning.
For our AR-Diffusion, the training loss decreases with fewer and less pronounced fluctuations.
This suggests better stability in the training process compared to Diffusion Forcing.

\section{Reconstruction Performance}
\begin{figure*}[t]
    \centering
    \includegraphics[width=1.0\linewidth]{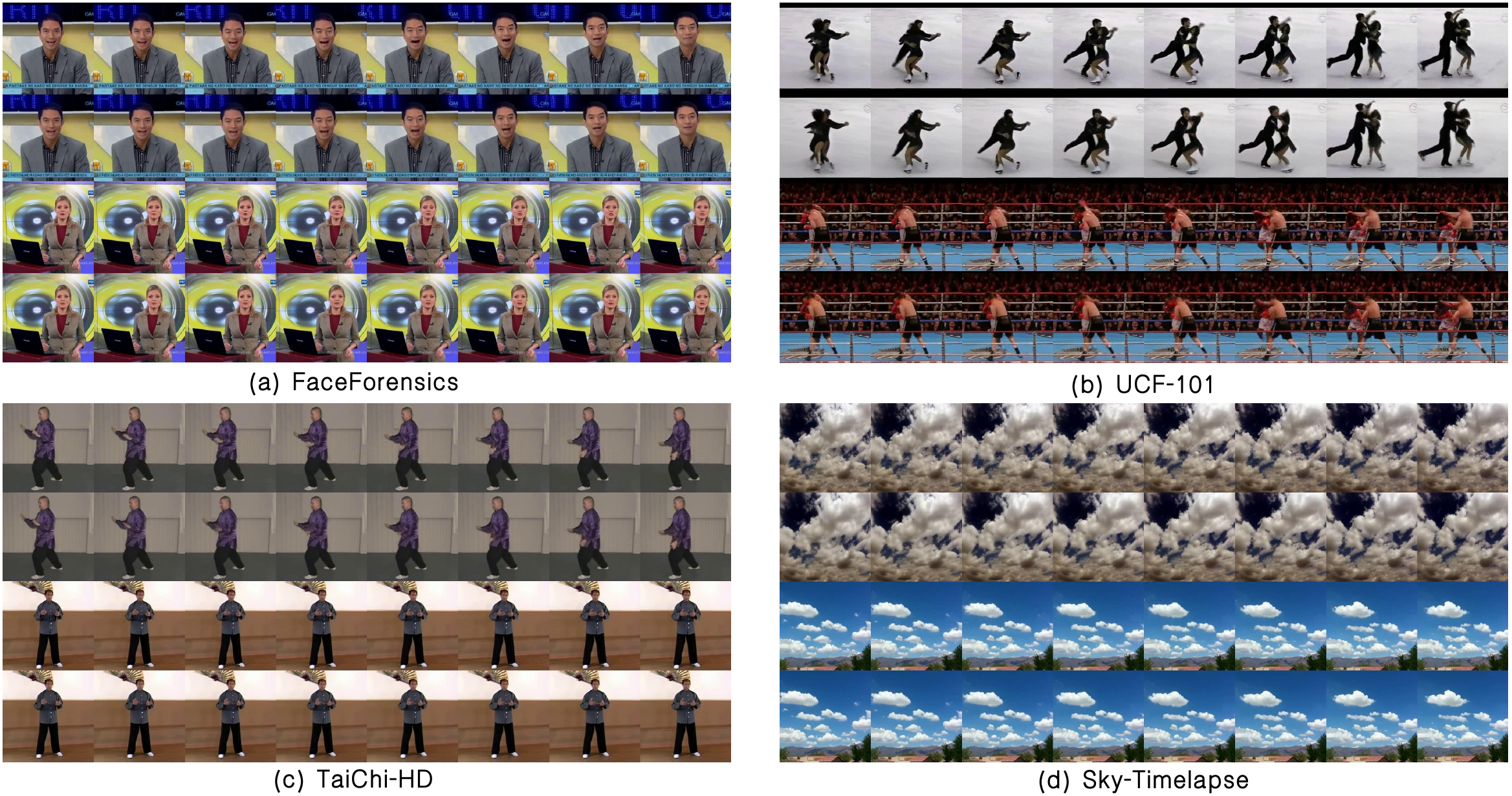}
    \caption{Real (first row) and reconstructed (second row) video frames using our AR-VAE on the (a) FaceForensics \cite{faceforensics}, (b) UCF-101 \cite{ucf101}, (c) TaiChi-HD \cite{siarohin2019first}, and (d) Sky-Timelapse \cite{xiong2018learning} datasets.}
    \label{fig:vae_recon}
\end{figure*}
In this section, we qualitatively analyze the performance of the Auto-Regressive Video Auto-Encoder (AR-VAE) on reconstructing video frames across different datasets: FaceForensics, UCF-101, TaiChi-HD, and Sky-Timelapse.
The results are reported in \cref{fig:vae_recon}.
On the FaceForensics dataset, the reconstructed frames closely resemble the real frames, maintaining the overall structure and details of the scenes. The colors and shapes are well-preserved, indicating that AR-VAE effectively captures the essential features of the video frames.
On UCF-101, the AR-VAE demonstrates strong performance in reconstructing dynamic actions, such as the movements of the individuals in the video. The reconstructed frames retain the motion and spatial details, ensuring temporal consistency and clarity.
On the TaiChi-HD dataset, the reconstructed frames accurately reflect the subject movements. 
The AR-VAE maintains the continuity and fluidity of the actions, preserving the intricate details and background elements.
On Sky-Timelapse, the AR-VAE effectively reconstructs the changing sky scenes, capturing the variations in cloud formations and lighting conditions. The reconstructed frames exhibit high fidelity to the real frames, maintaining the temporal progression and visual consistency.
Overall, the AR-VAE shows impressive reconstruction capabilities across different types of video content, preserving both spatial and temporal features with high accuracy.

\begin{table}[t]
    \centering
    \caption{
    Ablation study on VAE using 4 A800 for 100 hours w/o the fine-tune stage.
    }
    \label{tab:ablation}
    \scalebox{0.7}{
        \begin{tabularx}{1.4\linewidth}{ 
            p{0.32\linewidth} | X<{\centering} | X<{\centering} | X<{\centering}  X<{\centering} | X<{\centering}  X<{\centering} }
        \toprule
        VAE                 & Token   & Sample & \multicolumn{2}{c|}{Taichi-HD} & \multicolumn{2}{c}{Sky-Timelapse} \\ 
        Model               & Length  & Time & FVD$_{rec}$ & FVD$_{gen}$  &FVD$_{rec}$ & FVD$_{gen}$   \\
        \midrule
        AR-VAE (ours)       & 16x32    & 3.1s     & 18.1  & 83.9     & 15.3     & 79.6 \\
        OpenSora-VAE-v1.2   & 4x32x32  & 30.7s    & 41.2  & 785.4    & 11.3     & 643.8 \\
        \bottomrule
        \end{tabularx}
}
\end{table}

\section{Ablation study on AR-VAE}
We conduct ablation study on AR-VAE and report the results in Table \ref{tab:ablation_vae}.
FVD$_{gen}$ is obtained by training AR-Diffusion with different VAEs.
Inference time involves both generation and decoding processes.
We train AR-Diffusion with different VAE on 4 A800 GPUs for 100 hours.
As reported in Table \ref{tab:ablation}, current SOTA VAE, i.e. Open-Sora-VAE, utilizes 8 times more tokens to represent a video than our AR-VAE, resulting in a smaller batch size (2 vs 16),  slower training and infer speed, and much poorer generation performance.

\section{Efficiency Comparison}
As reported in Table \ref{tab:efficiency}, our method shows superior efficiency compared to others.
We use official codes\&ckpts.
Different from reported in paper, here we include both generation and decoding time.
\begin{table}[t]
    \centering
    \caption{Sampling a $F$-frame video on an A800. * denotes $128^2$ reso.}
    \label{tab:efficiency}
    \scalebox{0.8}{
        \begin{tabularx}{1.2\linewidth}{ 
            p{0.25\linewidth} | X<{\centering} X<{\centering} X<{\centering} X<{\centering} | X<{\centering} }
        \toprule
                    & Latte     & FIFO      & TATS      & VIDM      & Ours \\
        \midrule
        Max Mem.    & 14.7GB    & 8.5GB     & 4.1GB     & 35.5GB    & 8.6GB  \\
        F=16        & 6.8s      & 595.8s    & 14.8s     & 115.9s    & 3.1s \\
        F=128       & 52.0s     & 1639.8s   & 49.9s*    & 380.0s*   & 45.8s \\
        \bottomrule
        \end{tabularx}
    }
\end{table}

\section{Settings of Hyper Parameters}
The detailed settings of model hyper parameters are presented in Table \ref{tab:hyper}.

\begin{table}[h] \small
    \centering
    \caption{Hyper-parameters of the AR-VAE and the DiT backbone of AR-Diffusion.}
    \label{tab:hyper}
    \scalebox{1.0}{
    \begin{tabularx}{1.0\linewidth}{ p{0.5\linewidth} | X<{\centering} }
    \toprule
    \multicolumn{2}{c}{\cellcolor{gray!30}  AR-VAE} \\
    \midrule
    Token Length $L$        &   32 \\
    Token Dimension $D$     &   4     \\
    Model Width             &   1024    \\
    Num Layers              & 24    \\
    Num Heads               & 16    \\
    MLP Ratio               & 4.0   \\
    \midrule
    \multicolumn{2}{c}{\cellcolor{gray!30}  AR-Diffusion} \\
    \midrule
    Scale Factor            &  0.5 \\
    Hidden Size             & 1152  \\
    Depth                   & 28    \\
    Num Heads               & 16    \\
    $\beta$ Linear Start    & 0.0001    \\
    $\beta$ Linear End      & 0.002    \\
    Num Timesteps           & 1000  \\
    \bottomrule
    \end{tabularx}
    }
\end{table}

\section{Samples on Long Video Generation}

\begin{figure*}
    \centering
    \includegraphics[width=1.0\linewidth]{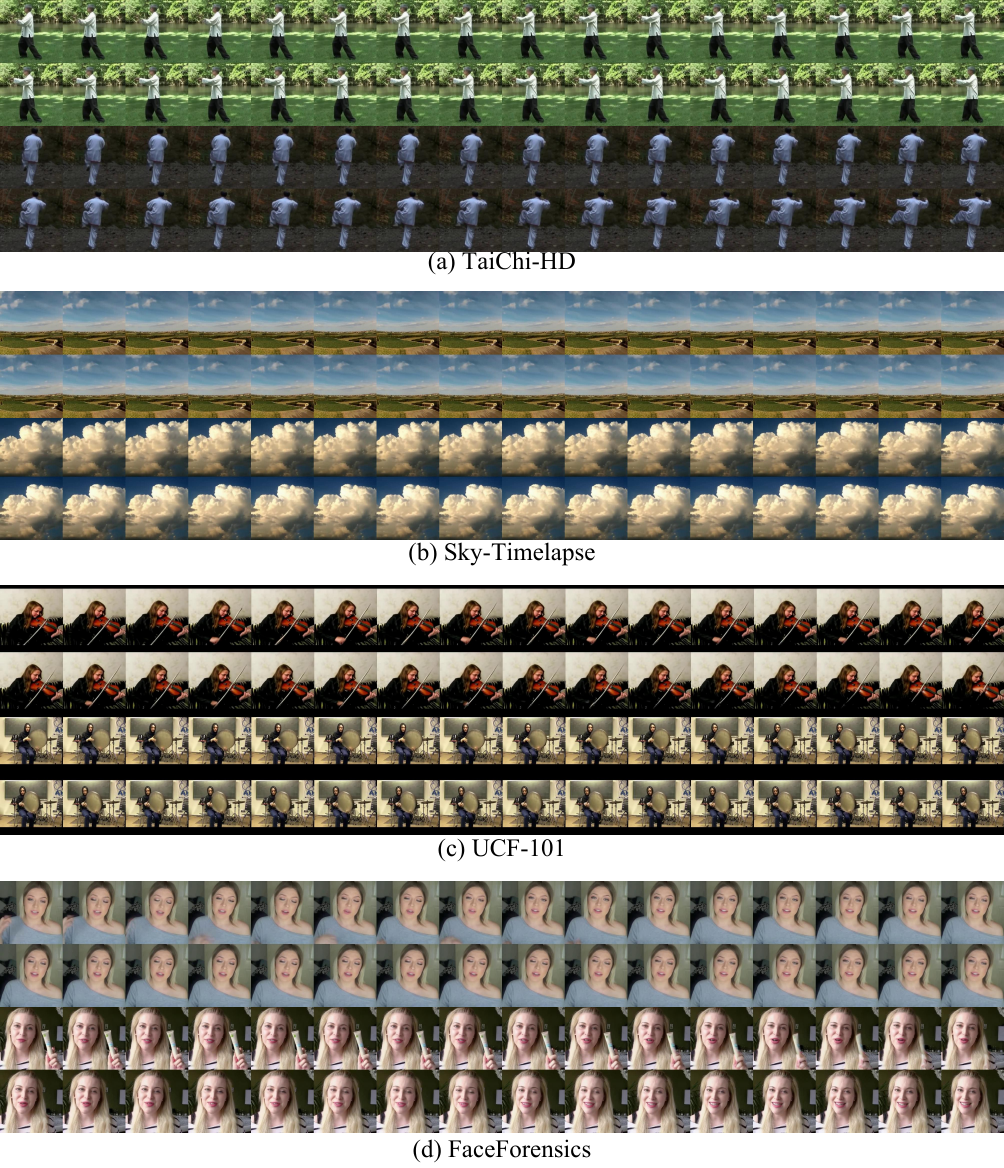}
    \caption{Generated 128-frame videos using our AR-Diffusion on four datasets. Each video is displayed with 4 skipped frames.}
    \label{fig:long}
\end{figure*}
In \cref{fig:long}, we present the generated 128-frame long videos by our AR-Diffusion on four diverse datasets: TaiChi-HD, Sky-Timelapse, UCF-101, and FaceForensics.
More displayable samples can be found in \url{https://anonymouss765.github.io/AR-Diffusion}.
The qualitative results of our AR-Diffusion model demonstrate its capability to generate visually realistic and temporally coherent video frames.
On the TaiChi-HD dataset, ehe generated frames exhibit smooth and natural transitions, capturing the fluid motion characteristic of Tai Chi exercises. The model maintains the consistency of the subject's movements and the background details, ensuring a coherent visual experience.
On the Sky-Timelapse dataset, ehe model effectively synthesizes the gradual changes in the sky, including cloud movements and lighting variations. The temporal coherence is well-preserved, with the transitions between frames appearing seamless and natural.
On the UCF-101 dataset, which includes various human actions, the AR-Diffusion model successfully generates frames that depict continuous and realistic motion. The actions are rendered with high fidelity, and the temporal progression of the activities is smooth and coherent.
On the FaceForensics dataset, the generated video frames show the model's ability to handle complex facial movements and expressions. The transitions between frames are smooth, and the facial details are consistently maintained, demonstrating the model's robustness in generating temporally coherent video sequences.
Overall, the AR-Diffusion model excels in producing high-quality video frames that are both visually realistic and temporally coherent, outperforming existing methods in handling diverse and challenging video generation tasks.

\end{document}